\documentclass[letterpaper]{article} 
\usepackage{aaai25}  
\usepackage{times}  
\usepackage{helvet}  
\usepackage{courier}  
\usepackage[hyphens]{url}  
\usepackage{graphicx} 
\usepackage{natbib}  
\usepackage{caption} 
\usepackage{algorithm}
\usepackage{algorithmic}

\usepackage{booktabs}       
\usepackage{amsfonts}       
\usepackage{nicefrac}       
\usepackage{microtype}      
\usepackage{xcolor}         
\usepackage{makecell}
\usepackage{booktabs}
\usepackage{dcolumn}
\usepackage{graphicx}
\usepackage{subcaption}
\usepackage{amsmath}
\usepackage{amsthm}
\theoremstyle{plain}
\newtheorem{theorem}{Theorem}

\theoremstyle{definition}
\newtheorem{definition}[theorem]{Definition}

\theoremstyle{remark}

\usepackage{thm-restate}
\usepackage{thmtools}
\usepackage{subcaption}
\definecolor{mygreen}{rgb}{0,0.8,0}

\usepackage{newfloat}
\usepackage{listings}
\DeclareCaptionStyle{ruled}{labelfont=normalfont,labelsep=colon,strut=off} 
\floatstyle{ruled}
\newfloat{listing}{tb}{lst}{}
\floatname{listing}{Listing}
\title{SORREL: Suboptimal-Demonstration-Guided \\ Reinforcement Learning for Learning to Branch}
\author{
    Shengyu Feng,
    Yiming Yang
}
\affiliations{
    Carnegie Mellon University\\

    \{shengyuf,yiming\}@cs.cmu.edu
}

\usepackage{bibentry}

\begin{document}

\maketitle

\begin{abstract}
Mixed Integer Linear Program (MILP) solvers are mostly built upon a Branch-and-Bound (B\&B) algorithm, where the efficiency of traditional solvers heavily depends on hand-crafted heuristics for branching. The past few years have witnessed the increasing popularity of data-driven approaches to automatically learn these heuristics. However, the success of these methods is highly dependent on the availability of high-quality demonstrations,  which requires either the development of near-optimal heuristics or a time-consuming sampling process. This paper averts this challenge by proposing \textit{Suboptimal-Demonstration-Guided Reinforcement Learning (SORREL)} for learning to branch. SORREL selectively learns from suboptimal demonstrations based on value estimation. It utilizes suboptimal demonstrations through both offline reinforcement learning on the demonstrations generated by suboptimal heuristics and self-imitation learning on past good experiences sampled by itself. Our experiments demonstrate its advanced performance in both branching quality and training efficiency over previous methods for various MILPs.
\end{abstract}

\section{Introduction}
Combinatorial optimization (CO) has been a fundamental challenge in computer science for decades, with a wide range of real-world applications, including
supply chain management, 
logistics optimization \citep{chopra2001strategy}, 
workforce scheduling \citep{ernst2004staff}, financial portfolios \citep{milpportfolio}, compiler optimization \citep{trofin2021mlgo,zheng2022alpa}, 
and more. 

Many of those CO problems can be formulated within a generic framework of Mixed Integer Linear Programs (MILPs), which is a central focus in algorithm development.  Traditional MILP solvers recursively apply a divide-and-conquer strategy to decompose a MILP into sub-problems with additional bounds on the variables in a tree search, namely Branch-and-Bound (B\&B) \citep{branchbound}, until an optimal solution is found. Off-the-shelf solvers of this kind include SCIP \citep{achterberg2009scip}, CPLEX \citep{cplex2009v12}, and Gurobi \citep{gurobi2021gurobi}.

In each iteration, the system solves the relaxed linear program (LP)  on a selected node (sub-problem) over the search tree and uses the LP solution (if it contains any fractional variable) to further
divide the current problem into two sub-problems. 
Such traditional solvers heavily rely on hand-crafted domain-specific heuristics for branching, which limits their capability to generalize across domains.

Recent machine learning research has offered new ways to solve MILPs by replacing the need for hand-crafted heuristics with automatically learned heuristics from training data \citep{Gasse2019Exact, Nair2020SolvingMI, scavuzzo2022learning, Parsonson_Laterre_Barrett_2023}.  As a representative example, \citet{Gasse2019Exact} formulated the MILP as a bipartite graph with variable nodes on the left and constraint nodes on the right, and trained a Graph Neural Network (GNN) to predict promising variables for branching in B\&B.  The follow-up works include the improvements of the GNN models for scaling up \citep{Nair2018Overcome, Gupta20hybrid} and improved solutions \citep{Zarpellon_Jo_Lodi_Bengio_2021, Huang2022Rank,zhang2024towards}. All of these models are trained via imitation learning (IL), and thus
have one limitation in common, i.e., their effectiveness relies on the availability of high-quality training data generated by near-optimal branching strategies
such as the
\textit{full strong branching} strategy \citep{Achterberg05branch}. However, there is no optimality guarantee for these heuristics on all types of MILPs, while developing a near-optimal branching heuristic on each type of MILPs poses a chicken-and-egg problem in learning.

To address the obstacle, some reinforcement learning (RL) methods have been proposed \citep{sun2020improving, scavuzzo2022learning, Qu2022AnIR, huang2023searching, Parsonson_Laterre_Barrett_2023}, which eliminate the dependency on the demonstrations from near-optimal heuristics by actively sampling training trajectories.  However, current RL-based methods still suffer from low sample efficiency in B\&B, and the training process often takes days to weeks even on easy MILPs that could be solved in seconds \citep{scavuzzo2022learning, Parsonson_Laterre_Barrett_2023}.

In essence, the practical applicability of existing methods has been significantly constrained by the trade-off between robustness to the demonstration quality and training efficiency. In this paper, we introduce a novel reinforcement learning method, namely \textit{\underline{S}ub\underline{o}ptimal-Demonst\underline{r}ation-Guided \underline{Re}inforcement \underline{L}earning (SORREL)}, to address this conflict by efficiently learning from suboptimal demonstrations. SORREL assimilates both (offline) IL- and (online) RL-based methods into a two-stage RL approach. In the first stage, SORREL trains an \textit{offline RL} branching agent on a static dataset with demonstrations precollected from existing heuristics. In addition to the standard supervised learning objective, the agent is trained to favor branching decisions that lead to higher estimated future returns, which minimizes the negative impact of low-quality demonstrations. In the second stage, SORREL uses (online) RL to further finetune the agent pretrained in the last stage, and we mainly develop two techniques to improve its training efficiency. Firstly, we develop and ground SORREL on a novel \textit{tree Markov Decision Process} \citep{Etheve2020FMSTS, scavuzzo2022learning}, which is both theoretically and empirically shown to be more general for modeling the variable selection process in B\&B. 
Secondly, SORREL utilizes a priority queue to track its best trajectories on each MILP instance and \textit{self-imitate} \citep{junhyuk2018sil} the past good branching decisions that achieve a higher return than expected.

Our empirical results demonstrate the clear advantage of SORREL in branching quality and generalization performance (trained on small instances and tested on larger ones). 
SORREL consistently outperforms representative baselines in reinforcement learning and imitation learning with access to the same suboptimal heuristics and achieves a comparable performance against the imitation learning method trained with high-quality demonstrations. Our analysis also reveals that SORREL achieves the best training efficiency among all neural methods. In short,  our findings suggest that SORREL holds promise as a potent neural MILP solver for practical applications.

\section{Preliminary}
\subsection{The B\&B Algorithm}
The Mixed Integer Linear Program (MILP) is defined by the linear object, linear constraints, and integrality constraints, which can be formally expressed as
\begin{equation}
    \min{\mathbf{c}^{\top}\mathbf{x}}, \ \text{s.t.} \ \mathbf{A}\mathbf{x}\leq \mathbf{b}, \ \mathbf{x}\in\mathbb{Z}^p\times\mathbb{R}^{n-p},
\end{equation}
with $\mathbf{c}\in\mathbb{R}^n$ the objective coefficient vector, $\mathbf{A}\in\mathbb{R}^{m\times n}$ the constraint coefficient matrix, $\mathbf{b}\in\mathbb{R}^{m}$ the constraint right-hand-side, and  $p\leq n$ the number of integer variables. When the integrality constraints are disregarded, we can obtain a linear program (LP) and solve it efficiently with algorithms like the simplex method. This process is known as linear programming relaxation, which yields a lower bound for the original problem since it is solved on a larger feasible region. If the LP relaxed solution $\mathbf{x}^{*}$ happens to be integral, then $\mathbf{x}^{*}$ is also guaranteed to be optimal for the original MILP. Otherwise, there must exist a set of variables $\mathcal{C}$ such that $\mathbf{x}^{*}_j$ is fractional for $j\in\mathcal{C}$. The B\&B algorithm then selects a variable from $\mathcal{C}$ to partition the problem into two child problems, with the additional constraint
\begin{equation}
    \mathbf{x}_j\leq\lfloor\mathbf{x}^{*}_j\rfloor \quad \text{or}\quad \mathbf{x}_j\geq\lceil\mathbf{x}^{*}_j\rceil.
\end{equation}
This partition process is known as variable selection or branching. With multiple sub-problems in hand, each time B\&B algorithm selects a sub-problem (node) to explore based on the heuristic known as the node selection policy. B\&B tracks the global primal/upper bound (lowest objective value for all feasible solutions) and dual/lower bound (highest objective value for all valid relaxations) throughout the solving and iterates through the aforementioned steps until the convergence of two bounds.

The quality of the branching policy has a high impact on the computational cost of B\&B. The branching policy needs to balance the size of the search tree and the computational cost for obtaining the branching decision. Among the current heuristics, \textit{full strong branching (FSB)} computes the actual change in the local dual-bound ($LDB$) by solving the resultant sub-problem for each fractional variable, which usually achieves a smaller search tree than competing methods \citep{ACHTERBERG200542}. However, the computational cost for obtaining the actual bound change itself is expensive and FSB becomes suboptimal when linear relaxation is not informative \citep{Gamrath2020AnEC}. Other common heuristics include \textit{pseudocost branching (PB)} that conducts a faster estimation of the change in $LDB$ at the cost of a larger search tree \citep{ACHTERBERG200542} and \textit{reliablility pseudocost branching (RPB)} (the default branching rule of SCIP) that uses FSB at the start of B\&B and switches to PB for the remaining steps \citep{ACHTERBERG200542}. 

\begin{figure*}[t]
    \centering
    \includegraphics[width=.9\linewidth]{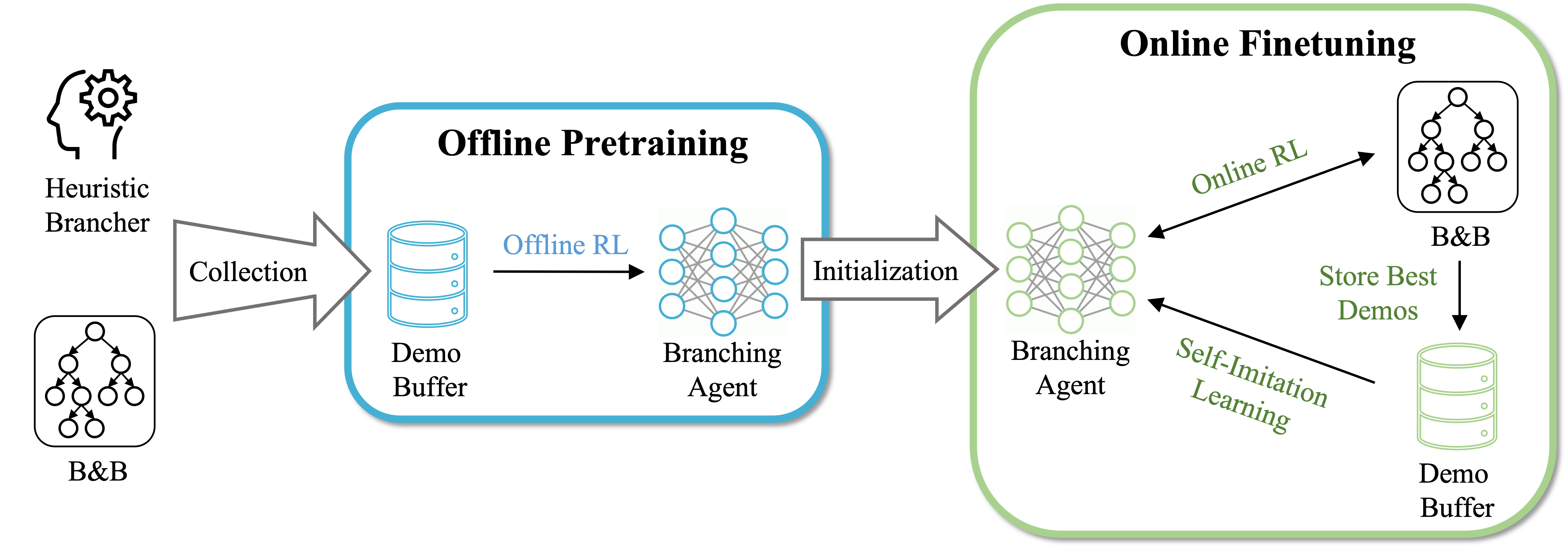} 
    \caption{Overview of  SORREL. SORREL first trains a branching agent via offline RL on demonstrations collected using existing heuristics, then finetunes the agent via online RL augmented by self-imitation learning from past good experiences.}
    \label{fig:sorrel}
\end{figure*}

\subsection{Reinforcement Learning}
\label{subsec:rl}
Reinforcement Learning (RL) is typically formalized through the framework of a Markov Decision Process (MDP) $(\mathcal{S}, \mathcal{A}, p_0, p_T, r, \gamma)$, with state space $\mathcal{S}$, action space $\mathcal{A}$, initial state distribution $p_0(s_0)$ and temporal transition dynamics $p_T(s'|s,a)$. Each time the agent receives a reward $r(s,a,s')\in \mathbb{R}$ by performing an action $a$ at state $s$ and reaching state $s'$. The return since state $s_t$ is a cumulative discounted reward weighted by the factor $\gamma\in[0,1)$, denoted as  $R_t=\sum_{t'=t}^{\infty}\gamma^{t'-t}r(s_{t'},a_{t'},s_{t'+1})$. RL aims to find a policy $\pi(\cdot|s)$ such that the total expected return $\mathbb{E}_{s_0\sim p_{0}(\cdot),a_t\sim\pi(\cdot|s_t), s_{t+1}\sim p_T(\cdot|s_t,a_t)}[R_0]$ is maximized. Each policy $\pi$ has an associated state-action value function $Q^{\pi}(s,a)=\mathbb{E}_{\pi}[R_t|s_t=s,a_t=a]$, corresponding to the expected return by following the policy $\pi$ after taking action $a$ in state $s$. Given any policy $\pi$, the state-action value function could be computed via the Bellman operator $\mathcal{T}^{\pi}$ 
\begin{equation}
    \mathcal{T}^{\pi}Q(s,a) = \mathbb{E}_{s',a'}[r(s,a,s')+\gamma Q(s',a')],
\end{equation}
which is a contraction \citep{bellman} with the unique fixed point $Q^{\pi}(s,a)$.
\section{Method}
\begin{figure}[b!]
\begin{subfigure}[b]{0.46\linewidth}
    \includegraphics[height=4.3cm,trim=0cm 0cm 0cm 0.5cm]{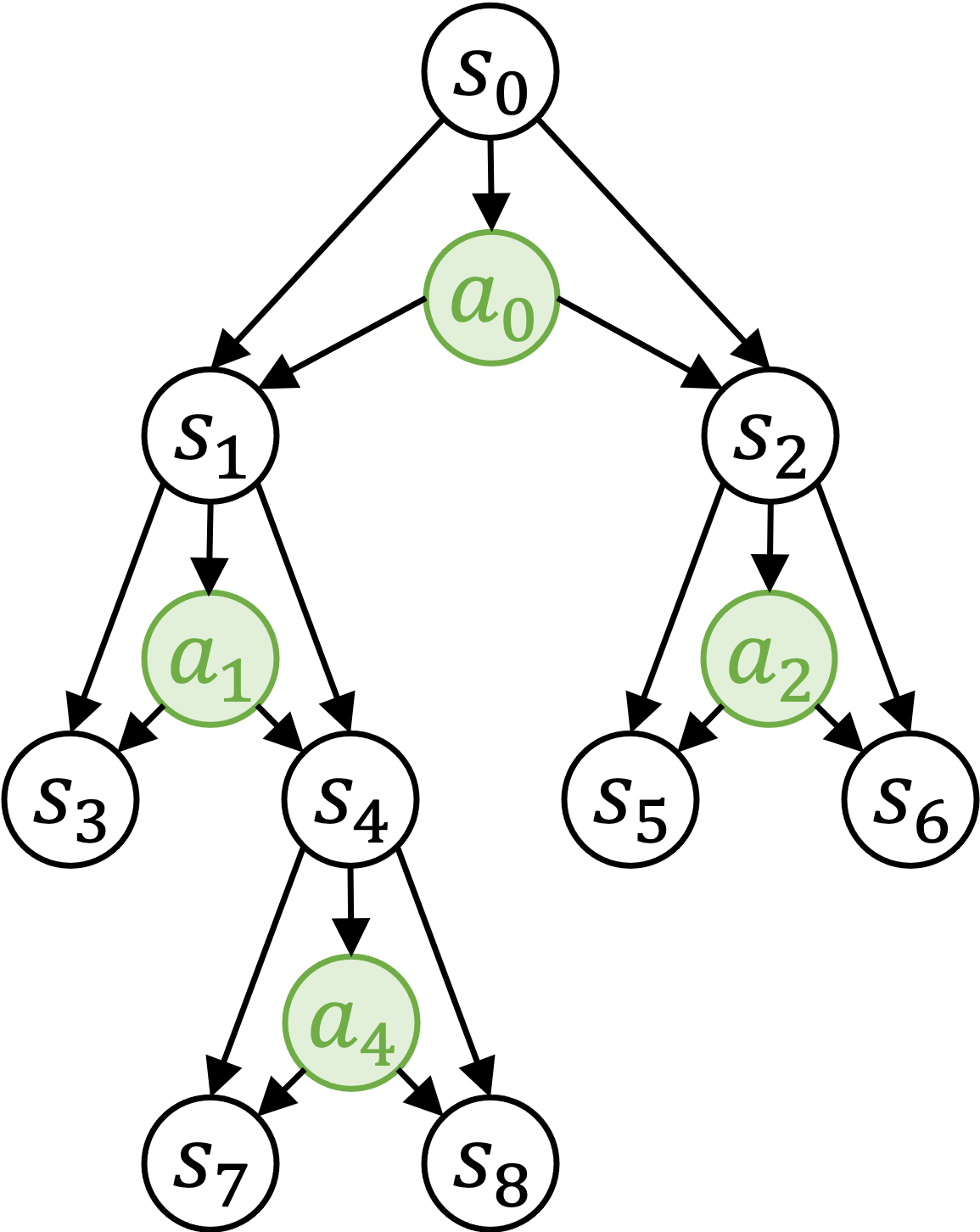} 
    \caption{The diagram of a tree MDP. Taking an action (green nodes) at each state (white nodes) yields two child states.}
    \label{fig:tree}
\end{subfigure}
\hfill
\begin{subfigure}[b]{0.46\linewidth}
    \includegraphics[height=4.3cm,trim=1.5cm 0cm 0cm 0.5cm]{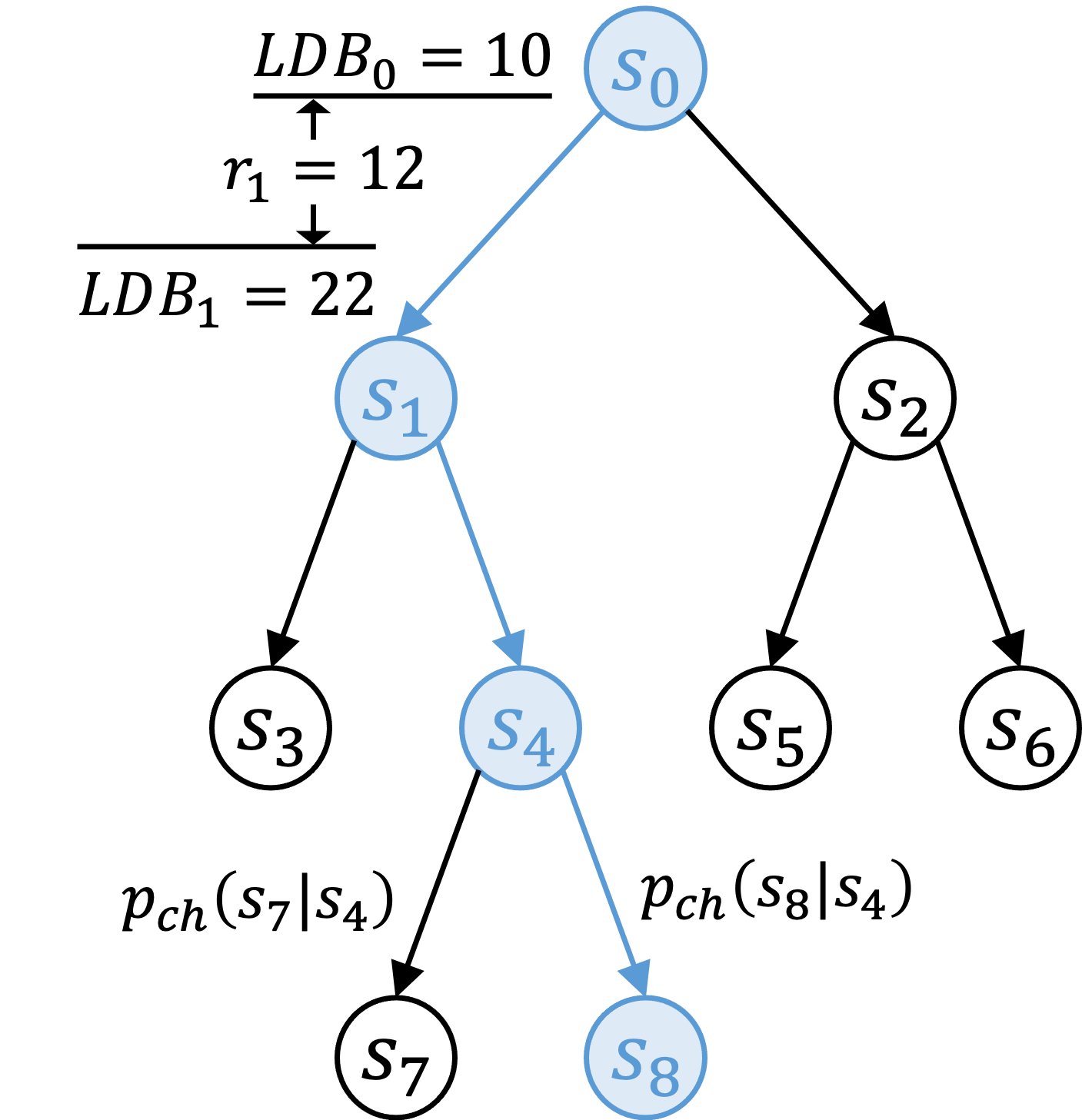} 
    \caption{Illustration of the random-walk-based return for $s_0$. The blue color highlights the sampled temporal trajectory.}
    \label{fig:return}
\end{subfigure}

\caption{Tree MDP \& its associated rewards.}
\end{figure}
Our approach, SORREL, is a two-stage RL method developed to achieve both robustness to the demonstration quality and training efficiency.
In SORREL, we use offline RL to achieve the robustness to the demonstration quality and develop multiple techniques to improve its training efficiency, including a generalized tree MDP and self-imitation learning. The overview of SORREL is shown in  Figure \ref{fig:sorrel}.

\subsection{Tree MDP for B\&B}
We start the introduction of SORREL with the RL formulation of branching. The variable selection process in B\&B could be modeled as an augmented MDP called the tree MDP \cite{Etheve2020FMSTS,scavuzzo2022learning}, as shown in Figure \ref{fig:tree}. Compared to standard MDPs, the tree MDP can simplify the credit assignment problem, that is, the problem of identifying the action responsible for a certain outcome, in B\&B. A tree MDP is defined as $(\mathcal{S}, \mathcal{A}, p_{0}, p_{ch}^-, p_{ch}^+, r, \gamma)$, with the major deviation from a standard MDP in its left and right child transition dynamics, $p_{ch}^-(s_{ch_i^-}|s_i,a_i)$ and $p_{ch}^+(s_{ch_i^+}|s_i,a_i)$, instead of the temporal transition dynamics $p_T(s'|s,a)$. In tree MDPs, taking an action $a_i$ in state $s_i$ would create two new states, left child $s_{ch_i^-}$ and right child $s_{ch_i^+}$, paired with rewards $r(s_i,a_i,s_{ch_i^-})$ and  $r(s_i,a_i,s_{ch_i^+})$ respectively. For simplicity, we use $r_{ch_i}$ as the short for $r(s_i,a_i,s_{ch_i})$ ($ch_i\in\{ch_i^-, ch_i^+\}$) in the following text.

The B\&B process modeled  by \citet{scavuzzo2022learning} can only be formulated as a tree MDP under certain conditions, e.g., a depth-first-search node selection strategy, we instead propose the following universal model for B\&B processes.
\begin{definition}
    In B\&B, define state $s_i=\textit{MILP}_i$, where $\textit{MILP}_i$ is the corresponding local sub-MILP, action $a_i$ as the selected variable for branching and reward $r(s_i,a_i, s_{ch_i})$ as the local-dual-bound improvement $\min_{ch} LDB_{ch}-LDB_{i}$.
    \label{def:rl}
\end{definition}
\begin{restatable}{proposition}{mdp}
\label{bnbmdp}
    Any B\&B process following Definition \ref{def:rl} can be formulated as a tree MDP.  
\end{restatable}
Our proposed model does not exactly follow the actual B\&B process but instead encourages local dual-bound improvement in the long run. Enhancing local dual-bounds itself is a frequently pursued goal in branching strategies \citep{achterberg2008constraint}, 
which also offers dense rewards compared to the size of the whole search tree.

Unlike in standard MDPs, the trajectory in tree MDPs consists of a binary tree, and there is no definite way to define the future return. In this work, we propose to define the return in tree MDPs based on a random walk process as illustrated in Figure \ref{fig:return}. For a (sub-)tree rooted at state $s_i$, we consider a random walk process starting at $s_i$ and each time it moves to a child state according to a distribution $p_{ch}$. This process will build a temporal trajectory as in standard MDPs, i.e., $(s_{i_0}, s_{i_1}, s_{i_2}, \cdots)$, where $i_0=i$, and $s_{i_{t+1}}$ is a child of $s_{i_t}$. The return for this temporal trajectory is simply the cumulative discounted reward $\sum_{t=0}\gamma^tr_{i_{t+1}}$ in standard MDPs. We then define the return for state $s_i$ in tree MDPs as the expected return over all possible temporal trajectories from the above random walk process, i.e.,
    $R_i=\mathbb{E}_{s_{i_{t+1}}\sim p_{ch}(\cdot|s_{i_t})}[\sum_{t=0}^{\infty}\gamma^tr_{i_{t+1}}|i_0=i]$.
Similarly, we can define the value function for policy $\pi$ as $Q^{\pi}(s_i,a_i)=\mathbb{E}_{\pi}[R_i]$ and the Bellman operator $\mathcal{T}_{tree}^{\pi}$ as
\begin{equation}
\label{eq:tree_bellman}
    \mathcal{T}_{tree}^{\pi}Q(s_i,a_i)=\mathbb{E}_{s_{ch_i},a_{ch_i}}[r_{ch_i}+\gamma Q(s_{ch_i}, a_{ch_i})].
\end{equation}
It can be proven that this new Bellman operator still inherits its property in standard MDPs.
\begin{restatable}{proposition}{bellman}
    The Bellman operator $\mathcal{T}_{tree}^{\pi}$ in Equation \ref{eq:tree_bellman} for tree MDPs is a $\gamma$-contraction operator in the $\mathcal{L}_{\infty}$ norm.
\end{restatable}

The above return definition could also be interpreted as the expected local dual-bound improvement in all leaf nodes against the root node, based on the stationary distribution. It thus offers us the flexibility to specify the preference among all leaf nodes, e.g., according to the node selection policy. 
\begin{restatable}{proposition}{randomwalk}
On a finite tree, for any distribution over the leaf nodes, there exists a set of $p_{ch}$ defined over all non-leaf nodes such that the stationary distribution of the above random walk is the same as the target distribution.
\end{restatable}

In this work, we let $p_{ch}(s_{ch_i}|s_i)=\kappa$ for $s_{ch_i}$ with the lower $LDB$ among child states, where $\kappa$ is a hyperparameter. Since $\gamma\in [0,1)$, maximizing the return would encourage the brancher to improve the dual bound as soon as possible, and thus reduce the size of the search tree as well. 

\subsection{Pretraining with Offline Reinforcement Learning}
Given the inefficiency in training a branching agent from scratch via online RL, SORREL first pretrains a branching agent with offline RL from the demonstrations collected using existing heuristic branchers. Here we denote the data buffer that stores the demonstrations as $\mathcal{D}_1=\{(s_i,a_i,s_{ch_i^-}, s_{ch_i^+})\}_{i=1}^N$. Instead of only imitating the branching decisions in the buffer, offline RL estimates the state-action value function and learns a policy that maximizes expected returns, thus achieving greater robustness in learning from suboptimal demonstrations compared to IL methods \citep{Kostrikov2021OfflineRL}.  

There are two networks in our offline RL algorithm, a policy network $\pi_{\phi}$ (actor) and a value network $Q_{\theta}$ (critic), as in the established actor-critic algorithm \citep{NIPS1999_6449f44a}. The value network aims to estimate the state-action value function for policy $\pi_{\phi}$ and is trained via the Bellman operator defined in Equation \ref{eq:tree_bellman},
\begin{equation}
\begin{split}
        \theta\leftarrow \arg\min_{\theta}&\mathbb{E}_{(s_i,a_i,s_{ch_i^-},s_{ch_i^+})\sim\mathcal{D}_1}[(Q_{\theta}(s_i,a_i)-\\
        &\mathbb{E}_{s_{ch_i},a_{ch_i}}[r_{ch_i}+\gamma Q_{\theta'}(s_{ch_i}, a_{ch_i})])^2],
\end{split}
\end{equation}
where $Q_{\theta'}$ is a slowly updated target value network used for a stable estimation of the target state-action value. The policy network is typically trained to maximize the expected state-action value, i.e., $\arg\max_{\phi}\mathbb{E}_{s_i,a\sim\pi_{\phi}(\cdot|s_i)}[Q(s_i,a)]$. However, such a training target usually suffers from the notoriously known \textit{distributional shift} problem in offline RL \citep{Kumar2019StabilizingOQ, Levine2020OfflineRL}. We address this issue by introducing the behavior cloning (BC) regularization in its training target, as proposed by \citet{fujimoto2021minimalist},
\begin{equation}
\begin{split}
       \phi\leftarrow \arg\max_{\phi}\mathbb{E}_{(s_i,a_i)\sim\mathcal{D}_1}&[\lambda\mathbb{E}_{a\sim\pi_{\phi}(\cdot|s_i)}[Q(s_i,a)]
       \\&+\log{\pi_{\phi}(a_i)}].
\end{split}
\end{equation}
Here, $\lambda$ controls the relative portion of the regularization term, whose value is chosen to be
\begin{equation}
    \lambda=\frac{\alpha}{\frac{1}{N}\sum_{(s_i,a_i)\sim\mathcal{D}_1}|Q(s_i,a_i)|},
\end{equation}
where $\alpha$ is a hyperparameter and $\frac{1}{N}\sum_{(s_i,a_i)\sim\mathcal{D}_1}|Q(s_i,a_i)|$ is estimated from the mini-batch.


\subsection{Finetuning with Self-Imitation Learning}
It is always challenging for an online RL agent to sample enough good trajectories in B\&B due to its large exploration space, in terms of both the large number of branching candidates and the deep search tree. This leads to a severe issue for standard RL algorithms in the sample efficiency, i.e., considerable time is spent in sampling and the agent cannot get effectively trained until high-quality trajectories are acquired. SORREL mitigates low sample efficiency in two aspects, tracking the best demonstrations on each MILP instance and learning from these suboptimal demonstrations through self-imitation learning (SIL) \citep{junhyuk2018sil}. In detail, SORREL finetunes the pretrained agent on a small number of MILP instances, each time it samples a subset of MILP instances for interaction, tracks the best few trajectories on each MILP instance, and stores them in buffer $\mathcal{D}_2=\{PQ_1, \cdots, PQ_M\}$. Here, $M$ is the number of MILP instances, and $PQ$ stands for the size-limit priority queue. Each $PQ$ stores the best few trajectories on the corresponding MILP instance, and each trajectory consists of tuples $(s_i,a_i, R_i)$ with state $s_i$, action $a_i$, and return $R_i$. SORREL identifies the good experiences in the buffer and trains the agent to imitate these branching decisions.

There are also two networks in our online RL algorithm, the policy $\pi_{\phi}$  initialized from the pretrained policy and the state value function $V_{\theta}$, which shares the same parameters with the pretrained state-action value function $Q_{\theta}$ but with a readout (pooling) function applied on top, e.g., $V_{\theta}(s_i)=\max_a Q_{\theta}(s_i,a)$. At each iteration, the agent samples a subset of MILP instances $\{I_k|k\}$ and collects the trajectory $\tau_k=\{(s_i,a_i,R_i)|i\}$ with $a_i\sim \pi_{\phi}(\cdot|s_i)$ on each instance. SORREL first performs the on-policy update with Proximal Policy Optimization (PPO) \citep{schulman2017proximal} 
\begin{align} 
\begin{split}
    \phi \leftarrow \arg\max_{\phi} &\mathbb{E}_{(s_i,a_i, R_i)\sim\bigcup\tau_k}[\min\{\frac{\pi_{\phi}(a_i|s_i)}{\pi_{\phi_{old}}(a_i|s_i)}A_i,\\
    & \quad clip(\frac{\pi_{\phi}(a_i|s_i)}{\pi_{\phi_{old}}(a_i|s_i)}, 1-\epsilon, 1+\epsilon)A_i\}],
\end{split}\\
    \theta\leftarrow \arg\min_{\theta}&\mathbb{E}_{(s_i,a_i,R_i)\sim\bigcup\tau_k}[(R_i-V_{\theta}(s_i))^2],
\end{align}
where $A_i=R_i-V_{\theta}(s_i)$ is the estimated advantage, and $\epsilon\in(0,1)$ is a hyperparameter of the clipping threshold.
 SORREL then updates the priority queues as $PQ_{k}\leftarrow PQ_{k} \cup \{\tau_{k}\}$ and the size of the search tree is used as the priority (smaller trees are prioritized). Finally, SORREL performs an off-policy update through SIL with the experiences sampled from $\mathcal{D}_2$, where the quality of the state-action pair $(s_i,a_i)$ is evaluated through $A_i$. The SIL updates can be written as
 \begin{align}
     \phi \leftarrow \arg\max_{\phi} & \mathbb{E}_{(s_i,a_i, R_i)\sim\mathcal{D}_2}[\log \pi_{\phi}(a_i|s_i)(A_i)_+],\\
     \theta  \leftarrow \arg\min_{\theta} &\mathbb{E}_{(s_i,a_i,R_i)\sim\mathcal{D}_2}[(R_i-V_{\theta}(s_i))_+^2].
 \end{align}
Here, $(\cdot)_+=\max(\cdot,0)$ filters out the good state-action pairs with a higher return than expected.

\section{Related Work}
Traditional MILP solvers rely on plenty of hand-crafted heuristics during their execution. Neural solvers therefore aim to improve these heuristics with deep learning methods \citep{BENGIO2021405}. Current neural solvers have successfully improved the performance of neural solvers by learning the heuristics in variable selection (branching) \citep{Gasse2019Exact, Gupta20hybrid, Nair2020SolvingMI, Zarpellon_Jo_Lodi_Bengio_2021, scavuzzo2022learning, Huang2022Rank}, node selection \citep{He2014Learn, Song2018LearningTS}, cutting plane selection \citep{Tang2020Cut, Paulus22aCut, Turner2023Cut, wang2023learning}, large neighborhood search \citep{sun2020improving, wu2021learning, Sonnerat2021LearningAL, huang2023searching}, diving \citep{Nair2020SolvingMI, yoon2022confidence, han2023a, paulus2023learning} and primal heuristics selection \citep{ijcai2017p92, Hendel2019Adaptive, Chmiela2021LearningTS}. Our work studies the variable selection heuristic, which receives the most attention in neural solvers.

  \citet{Khalil2016Rank, alvarez2017ml, Hansknecht2018CutsPH} are the earliest works to use statistical learning for the branching heuristic. They use an IL method to first collect an offline dataset with FSB and then treat the learning as either a ranking \citep{Khalil2016Rank, Hansknecht2018CutsPH} or regression problem \citep{alvarez2017ml}. With the advent of GNNs, \citet{Gasse2019Exact} transform each MILP instance into a bipartite graph and train a GNN classifier to imitate the choice of FSB. This work lays out the basic model architecture for future neural solvers on variable selection. To extend this GNN-based neural solver to larger instances, \citet{Nair2020SolvingMI} adopt a more efficient batch linear programming solver based on the alternating direction method of multipliers, and \citet{Gupta20hybrid} introduce a hybrid model, with a GNN applied only at the root node and using a light classifier for future predictions. An even more powerful IL model is proposed by \citet{gupta2022lookback} to take into account the lookback phenomenon in FSB.
  Recently, \citet{Etheve2020FMSTS, scavuzzo2022learning, sorokin2023treedqn} formulate B\&B as a tree MDP to mitigate the credit assignment problem in RL for branching methods, while \citet{Parsonson_Laterre_Barrett_2023} develop an enhanced RL method to learn from retrospective trajectories. There are also some recent works \citep{Huang2022Rank, Qu2022AnIR, zhang2022deep} using a similar hybrid framework of offline and online RL in learning to branch, but the suboptimality is not in the research scope of their study. A parallel work to ours at the same time is \citet{zhang2024towards}, which is also motivated by the potential suboptimality in demonstrations. However, they still focus on the demonstrations generated by FSB, and do not quantitatively study the suboptimality of the demonstrations or consider the training efficiency.

\section{Experiment}
\subsection{Experimental Setup}
\paragraph{Baselines.} We compare our method to two classical branching heuristics, full strong branching (FSB) and reliability pseudocost branching (RPB), and three neural methods, including the online RL methods TreeREINFORCE \citep{scavuzzo2022learning} and TreeDQN \citep{sorokin2023treedqn}, and IL method GCNN \citep{Gasse2019Exact}. Besides, we also design a vanilla hybrid branching (VHB) heuristic, which adopts FSB with probability 0.05 at each decision step and uses the pseudocost branching otherwise. VHB serves as the suboptimal heuristic for our demonstration collection across the experiments, and the GCNN method trained on these demonstrations is denoted as GCNN-VHB. For all neural methods, we standardize the GNN architecture as in \citet{Gasse2019Exact}.


\paragraph{Benchmarks.} We evaluate all methods on five commonly used MILP benchmarks for NP-hard problems, including \textit{Set Covering (SC)}, \textit{Maximal Independent Set (MIS)}, \textit{Combinatorial Auction (CA)}, \textit{Capacitated Facility Location (CFL)} and \textit{Multiple Knapsack (MK)}. We follow the instance generation process in \citet{Gasse2019Exact} to generate 10,000 MILP instances for training, 2,000 instances for validation, and 100 instances for testing, with the same distribution. For consistency, 20 validation instances and 100 training instances are held for  validation and online finetuning of SORREL. We also generate 20 larger testing instances from a transfer distribution to evaluate the generalization performance. 

We collect the suboptimal demonstrations for training by running VHB on randomly sampled training instances to generate the B\&B search tree until 100,000 transitions are collected. The same process is repeated to create the validation set with 20,000 transitions for GCNN.  

\begin{table*}
    \centering
    \begin{tabular}{l | D{,}{\;\pm\;}{5} D{,}{\;\pm\;}{5} D{,}{\;\pm\;}{5} D{,}{\;\pm\;}{5} D{,}{\;\pm\;}{5}}
        & \multicolumn{1}{c}{SC} & \multicolumn{1}{c}{MIS} & \multicolumn{1}{c}{CA} & \multicolumn{1}{c}{CFL} &  \multicolumn{1}{c}{MK} \\    
    \toprule
    Model & \multicolumn{1}{c}{Time (s) $\downarrow$} & \multicolumn{1}{c}{Time (s) $\downarrow$}  & \multicolumn{1}{c}{Time (s) $\downarrow$} & \multicolumn{1}{c}{Time(s) $\downarrow$} & \multicolumn{1}{c}{Time (s) $\downarrow$} \\
     \midrule
     FSB & 6.06 , 18.2\% & 61.49 , 72.0\% & 54.31 , 15.7\% & 96.64 , 66.9\% & 20.89 , 295.4\%\\
    RPB & 2.09 , 7.5\% & 5.59 , 15.6\% & \textbf{3.60} , \textbf{8.8\%} & \textcolor{mygreen}{29.28} , \textcolor{mygreen}{27.8\%} & \textbf{0.64} , \textbf{114.9\%}\\
    VHB & 2.71 , 33.9\% & 15.18 , 90.2\% & 12.21 , 38.5\% & 33.60 , 36.1\% & 6.53 , 654.3\% \\
    TreeREINFORCE & 3.10 , 23.8\% & \textcolor{mygreen}{4.72} , \textcolor{mygreen}{334.1\%} & 7.00 , 20.6\% & 35.49 , 39.7\% & 3.95 , 105.1\% \\
    TreeDQN & 2.73 , 22.4\% & 7.41 , 769.1\% & 2083.69 , 8.2\% & 33.78 , 37.0\% & 2.85 , 83.1\%\\
    GCNN-VHB & \textcolor{mygreen}{2.05} , \textcolor{mygreen}{8.8\%} & 5.51 , 28.8\% & 8.97 , 17.6\% & 30.02 , 39.1\% & 10.52 , 252.4\%\\
     SORREL (Ours) & \textbf{1.95} , \textbf{10.7\%} & \textbf{4.46} , \textbf{18.2\%} & \textcolor{mygreen}{4.40} , \textcolor{mygreen}{14.5\%} & \textbf{27.81} , \textbf{35.0\%} & \textcolor{mygreen}{2.54} , \textcolor{mygreen}{76.1\%}\\
     \bottomrule
    \toprule
 Model & \multicolumn{1}{c}{\# Nodes $\downarrow$} & \multicolumn{1}{c}{\# Nodes $\downarrow$}  & \multicolumn{1}{c}{\# Nodes $\downarrow$} & \multicolumn{1}{c}{\# Nodes $\downarrow$} & \multicolumn{1}{c}{\# Nodes $\downarrow$}\\
     \midrule
     FSB & 24 , 13.2\% & 36 , 84.5\% & 195 , 17.2\% & 173 , 44.5\% & 522 , 831.7\% \\
     RPB & 9 , 61.6\% & 23 , 177.8\% & 85 , 88.7\% & 94 , 86.3\% & 356 , 627.0\% \\
     VHB & 42 , 43.0\% & 115 , 200.4\% & 777 , 31.6\% & 257 , 37.5\% & 563 , 1466.3\% \\
     \midrule
     TreeREINFORCE & 75 , 86.6\% & \textcolor{mygreen}{72} , \textcolor{mygreen}{1409.0\%} & \textcolor{mygreen}{367} , \textcolor{mygreen}{21.6\%} & 322 , 48.4\% & 183 , 244.4\%\\
    TreeDQN &  68 , 85.2\% & 115 , 2584.1\% & / , / & 331 , 51.2\% & \textcolor{mygreen}{159} , \textcolor{mygreen}{150.3\%}\\
     GCNN-VHB & \textcolor{mygreen}{37} , \textcolor{mygreen}{24.8\%} & 93 , 189.3\% & 582 , 22.8\% & \textcolor{mygreen}{263} , \textcolor{mygreen}{44.0\%} & 458 , 675.2\% \\
    SORREL (Ours) & \textbf{32} , \textbf{18.7\%} & \textbf{44} , \textbf{159.4\%} & \textbf{253} , \textbf{18.0\%} & \textbf{239} , \textbf{43.6\%} & \textbf{142} , \textbf{192.2\%}\\
     \bottomrule
    \end{tabular}
    \caption{Comparative results in the solving time and the size of the search tree on the \textit{standard testing instances}, which are of the same size as training instances.  We bold the best results and color the second-best in green on each dataset for each metric. Only \textit{neural methods} are compared in the number of nodes. '/' means not comparable due to overwhelming unfinished runs.}
    \label{tab:standard}
\end{table*}

\begin{table*}
    \centering
    \begin{tabular}{l | D{,}{\;\pm\;}{5} D{,}{\;\pm\;}{5} D{,}{\;\pm\;}{5} D{,}{\;\pm\;}{5} D{,}{\;\pm\;}{5}}
        & \multicolumn{1}{c}{SC} & \multicolumn{1}{c}{MIS} & \multicolumn{1}{c}{CA} & \multicolumn{1}{c}{CFL} &  \multicolumn{1}{c}{MK} \\    
    \toprule
    Model & \multicolumn{1}{c}{Time (s) $\downarrow$} & \multicolumn{1}{c}{Time (s) $\downarrow$}  & \multicolumn{1}{c}{Time(s) $\downarrow$} & \multicolumn{1}{c}{Time (s) $\downarrow$} & \multicolumn{1}{c}{Time (s) $\downarrow$} \\
     \midrule
     FSB & 145.67 , 13.7\% & 2668.69 , 6.7\% & 3533.74 , 1.5\% & 413.02 , 42.0\% & 137.47 , 149.8\%\\
    RPB & 14.04 , 12.3\% & \textcolor{mygreen}{91.35} , \textcolor{mygreen}{56.7\%} & \textbf{71.18} , \textbf{7.8\%} & 111.39 , 17.8\% & \textbf{2.29} , \textbf{107.4\%}\\
    VHB & 21.35 , 25.8\% & 1902.31 , 13.8\% & 1680.47 , 15.4\% & 169.79 , 25.1\% & 17.89 , 939.7\% \\
    TreeREINFORCE & 55.08 , 54.7\% & 115.20 , 354.6\% & 874.82 , 31.8\% & 193.65 , 38.3\% & 10.20 , 112.0\%\\
    TreeDQN &  41.68 , 54.1\% & 404.05 , 133.7\% & 3600, 0.0\% & 111.81 , 15.0\% & 9.36 , 47.2\% \\
    GCNN-VHB & \textcolor{mygreen}{13.00} , \textcolor{mygreen}{5.9}\% & 424.53 , 53.9\% & 1492.86 , 20.6\% & \textcolor{mygreen}{108.3} , \textcolor{mygreen}{21.0\%} & 16.18 , 1031.5\%\\
     SORREL (Ours) & \textbf{12.88} , \textbf{7.0\%} &  \textbf{62.41} , \textbf{56.8\%} & \textcolor{mygreen}{366.49} , \textcolor{mygreen}{29.6\%} &\textbf{95.84} , \textbf{14.6\%}& \textcolor{mygreen}{8.98} , \textcolor{mygreen}{142.9\%}\\
     \bottomrule
    \toprule
 Model & \multicolumn{1}{c}{\# Nodes $\downarrow$} & \multicolumn{1}{c}{\# Nodes $\downarrow$}  & \multicolumn{1}{c}{\# Nodes $\downarrow$} & \multicolumn{1}{c}{\# Nodes $\downarrow$} & \multicolumn{1}{c}{\# Nodes $\downarrow$}\\
     \midrule
     FSB & 290 , 6.8\% & 161 , 27.9\% & 1402 , 15.6\% & 215 , 10.9\% & 679 , 1199.9\% \\
     RPB & 258 , 38.5\% & 2396 , 161.2\% & 11347 , 11.8\% & 128 , 36.1\% & 707 , 438.4\% \\
     VHB & 618 , 17.6\% & 3423 , 32.4\% & 20863 , 20.7\% & 315 , 18.9\% & 1187 , 1231.7\% \\
     \midrule
     TreeREINFORCE & 2411 , 95.2\% & \textcolor{mygreen}{2565} , \textcolor{mygreen}{714.1\%} & \textcolor{mygreen}{22284} , \textcolor{mygreen}{24.3\%} & 433 , 14.0\% & 536 , 539.2\% \\
     TreeDQN & 1915 , 102.3\% & 8836 , 127.0\% & /, /  & 451 , 14.6\% & \textcolor{mygreen}{480} , \textcolor{mygreen}{122.1\%} \\
     GCNN-VHB & \textcolor{mygreen}{509} , \textcolor{mygreen}{10.2\%} & 8511 , 67.6\% & 29748 , 21.1\% & \textcolor{mygreen}{378} , \textcolor{mygreen}{16.2\%} & 802 , 3264.9\% \\
    SORREL (Ours) & \textbf{485} , \textbf{10.3\%} & \textbf{1707} , \textbf{172.9\%} & \textbf{14520} , \textbf{37.0\%} & \textbf{324} , \textbf{16.1\%} & \textbf{408} , \textbf{595.3\%}\\
     \bottomrule
    \end{tabular}
    \caption{Comparative results in the solving time and the size of the search tree on the \textit{transfer testing instances}, which are larger than the training instances.  We bold the best results and color the second-best in green on each dataset for each metric. Only \textit{neural methods} are compared in the number of nodes. '/' means not comparable due to overwhelming unfinished runs.}
    \label{tab:transfer}
\end{table*}
\begin{table*}
\centering
\begin{tabular}{l | D{,}{\;\pm\;}{5} D{,}{\;\pm\;}{5} D{,}{\;\pm\;}{5} D{,}{\;\pm\;}{5} D{,}{\;\pm\;}{5}}
        & \multicolumn{1}{c}{SC} & \multicolumn{1}{c}{MIS} & \multicolumn{1}{c}{CA} & \multicolumn{1}{c}{CFL} & \multicolumn{1}{c}{MK} \\      
    \toprule
 Model & \multicolumn{1}{c}{\# Nodes $\downarrow$} & \multicolumn{1}{c}{\# Nodes $\downarrow$}  & \multicolumn{1}{c}{\# Nodes $\downarrow$} & \multicolumn{1}{c}{\# Nodes $\downarrow$} & \multicolumn{1}{c}{\# Nodes $\downarrow$}\\
     \midrule
    GCNN-FSB & 28 , 15.2\% & 37 , 81.7\% & 220 , 16.4\% & 218 , 43.4\% & 305 , 487.7\%\\
    \midrule
    GCNN-VHB &  37 , 24.8\% & 93 , 189.3\% & 582 , 22.8\% & 263 , 44.0\% & 458 , 675.2\%  \\
    SORRE-offline& 33 , 20.3\% & 47 , 169.9\% & 261 , 17.9\% & 241 , 40.8\% & 234 , 313.8\%\\
    SORREL  & \textbf{32} , \textbf{18.7\%} & \textbf{44} , \textbf{159.4\%} & \textbf{253} , \textbf{18.0\%} & \textbf{239} , \textbf{43.6\%} & \textbf{142} , \textbf{192.2\%}\\
     \bottomrule
    \end{tabular}
        \caption{Ablation study on proposed components in SORREL. The best results are in bold.}
    \label{tab:ablate}
\end{table*}

\paragraph{Metrics.} We use the solution time and the size of the B\&B search tree (measured by its number of nodes) as evaluation metrics. The former is a universal metric to compare both neural methods and traditional heuristics, while the latter is more straightforward for comparison when the decision time is the same, e.g., the neural methods. Five different random seeds are used during the evaluation, which leads to $100\times 5=500$ runs on standard testing instances and $20\times5=100$ runs on transfer testing instances. We report the 1-shifted geometric mean of solving time and the 10-shifted geometric mean of the number of nodes for all runs, following the conventions in \citet{inbook}. The average per-instance standard deviation is also attached.

\subsection{Main Results}  All our results are measured using SCIP 7.0.1 as the backend solver, employing a 1-hour time limit. Notably, we always keep the results from the hard instances and the solving is not interrupted only if the 1-hour time limit is reached. Although it leads to a higher standard deviation compared to the values reported in previous studies, it offers a new perspective on the model's performance consistency. 

Table \ref{tab:standard} and  \ref{tab:transfer} present the comparative results evaluated on the standard testing instances and transfer testing instances, respectively. SORREL consistently outperforms all neural methods in both the solving time and the search tree size in each table. Although SORREL lags behind RPB in solving time on the CA and MK datasets, it still appears as the fastest solver among all methods in 3 out of 5 datasets. SORREL demonstrates an impressive capability to learn from suboptimal demonstrations, achieving overall improvement over the behavior policy, VHB. In contrast, GCNN-VHB still exhibits inferior performance from time to time, for example, on CFL. Remarkably, SORREL outperforms the state-of-the-art method on the MK dataset, TreeDQN, by achieving a notably smaller search tree. Since linear relaxation is uninformative for MK, FSB also becomes suboptimal in this problem. Consequently, RL-based methods, such as TreeREINFORCE and TreeDQN, typically outperform their IL-based counterparts. However, by selectively learning good branching decisions from suboptimal demonstrations, SORREL can still benefit from these suboptimal demonstrations and achieve even better performance. Finally, an evident limitation of previous RL methods lies in their poor performance consistency. Higher variances are observed in most results for TreeREINFROCE and TreeDQN, while SORREL achieves more stable performance in general. The consistent improvement over neural baselines implies the generalizability of SORREL across various MILPs.

\subsection{Ablation Study}
In our ablation study, we first examine the efficacy of our offline RL algorithm by evaluating SORREL with offline RL only, denoted as SORREL-offline, in standard testing. The results are shown in Table \ref{tab:ablate} in terms of the number of nodes. We compare it to the full SORREL and GCNN-VHB. The performance of GCNN trained on the demonstrations from FSB, denoted as GCNN-FSB, is also included for reference. It can be seen that SORREL-offline has already achieved substantial improvements against GCNN-VHB across all datasets. Its ability to overcome suboptimality in demonstrations is especially impressive in the MK dataset, where SORREL can outperform GCNN-FSB without any online training as needed in previous RL methods. 
Online finetuning further boosts the performance of SORREL as suggested by Table \ref{tab:ablate}. There is a significant improvement in the MK dataset, and the performance gap towards GCNN-FSB is also notably reduced. Therefore, these results indicate that SORREL can benefit from both offline pretraining and online finetuning.

\begin{figure}[H]
    \centering    \includegraphics[width=\linewidth, trim=1cm 0cm 1cm 0cm,]{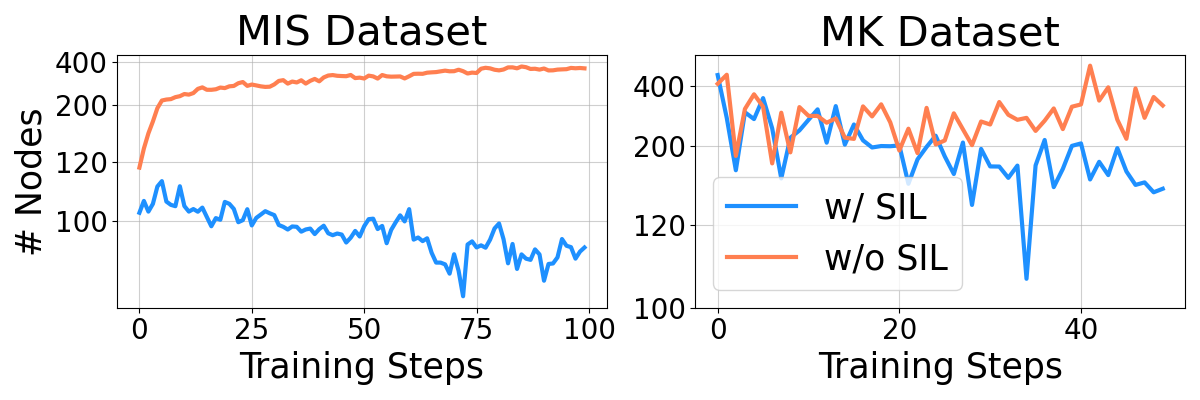} 
    \caption{Effect of SIL on the sample efficiency. Training dynamics in tree sizes are shown. The $y$-axis is in log-scale.}
    \label{fig:curve}
\end{figure}

Besides, we evaluate the impact of SIL on sample efficiency during online finetuning. In Figure \ref{fig:curve}, we use MIS and MK datasets to track the evolution of the validation B\&B tree sizes throughout training. When SIL is removed, our online RL algorithm is reduced to a standard PPO algorithm run on tree MDPs, similar to TreeREINFORCE. It can be seen that the model would quickly get stuck in a bad parameter area and suffer from high variances. Instead, SIL drives deep exploration and achieves stable improvement and efficient training, which is critical to the sample efficiency of SORREL. 
\begin{figure}[H]
        \centering
    \includegraphics[width=\linewidth]{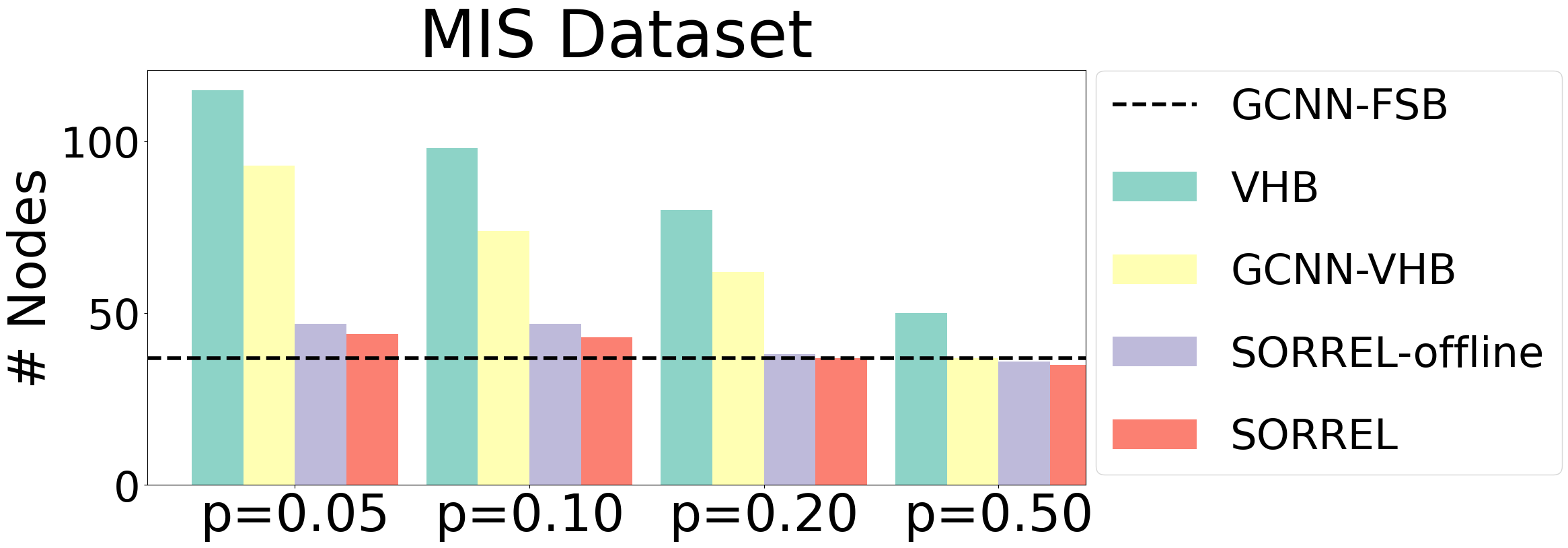} 
    \caption{Comparative results on the MIS datset under different levels of suboptimality (varying $p$ in VHB).}
    \label{fig:var_p}
\end{figure}

\subsection{Performance vs. Levels of Suboptimality}
In our main experiments, we simulate the situation where only suboptimal heuristics are available for learning / data collection. The VHB we used, with 95\% PB and 5\% FSB, is a synthetic suboptimal heuristic assumed to be the best available in that situation. Here we study the performance change of GCNN and SORREL when they are learning from heuristics with different levels of suboptimality. 

We use the MIS dataset as an example and vary the frequency of FSB in VHB to generate different learning heuristics.  Denote the frequency of using FSB in VHB as $p$, we choose $p$ from $\{0.05, 0.1, 0.2, 0.5\}$ and empirically a larger $p$ will lead to a better branching policy. The comparative results are visualized in Figure \ref{fig:var_p}.

It can be seen that as $p$ increases, the improved quality of VHB also increases the performance of GCNN-VHB and SORREL. However, we also have the following observations. (1) The relative performance, SORREL $>$ SORREL-offline $>$ GCNN-VHB, always holds whenever they are learning from the same heuristic (the same $p$). This verifies the consistent effectiveness of the main components in SORREL. 
(2) GCNN-VHB is more susceptible to the suboptimality in the learning heuristic than SORREL. (3) SORREL can outperform GCNN-FSB when we improve the quality of the learning heuristic (but still clearly worse than FSB).

\section{Conclusion \& Limitation}

In this paper, we propose a novel RL approach SORREL for neural branching in mixed linear integer programming. 
SORREL tackles the limitations of previous neural branching algorithms in their reliance on near-optimal heuristics or time-intensive sampling for high-quality training data. It efficiently learns from suboptimal demonstrations and outperforms previous IL- and RL-based neural branching methods in both branching quality and training efficiency. SORREL thus exhibits strong potential to generalize neural MILP solvers to more challenging problems. 

One limitation of SORREL lies in its inability to benefit from a longer RL training time due to the conservativeness of SIL. The balance of exploration and exploitation thus remains a meaningful topic in learning to branch. Design alternatives, including the reward function, returns, and priority queues, are still open for study. Future works can also extend SORREL to the learning of other heuristics in combinatorial optimization problems.

\bibliography{aaai25}

\end{document}